\documentclass[letterpaper, 10 pt, conference]{ieeeconf}  

\IEEEoverridecommandlockouts                              

\usepackage{amsmath}
\usepackage{amssymb}
\usepackage{cite}
\usepackage{graphicx}
\usepackage{tabularx}
\usepackage[table]{xcolor}
\usepackage{placeins}
\usepackage{float}
\definecolor{oursblue}{RGB}{229,241,255}

\title{\LARGE \bf
PACE: Stage-Consistent Long-Horizon Robot Manipulation via Progress-Aligned Context for Execution
}

\author{
Yenan Chen$^{1,\dagger}$,
Junjie Shi$^{1}$,
Lu Chen$^{1}$,
Zhongxiang Zhou$^{1}$,
and Rong Xiong$^{1,*}$%
\thanks{$^{1}$All authors are with the College of Control Science and Engineering,
Zhejiang University, Hangzhou 310027, China.}%
\thanks{$^{\dagger}$Contact: Yenan Chen
(\texttt{12532108@zju.edu.cn}).}%
\thanks{$^{*}$Corresponding author: Rong Xiong
(\texttt{rxiong@zju.edu.cn}).}%
}

\begin{document}

\maketitle
\thispagestyle{empty}
\pagestyle{empty}

\begin{abstract}

Demonstration-conditioned policies provide a natural interface for specifying robot behavior, yet long-horizon manipulation remains difficult when visually similar states recur across different stages or when demonstrations and executions proceed at different speeds. We identify the resulting failure mode as \emph{stage confusion} and introduce Progress-Aligned Context for Execution (PACE), a stateful method that continually reinterprets a complete demonstration according to realized execution progress. PACE compresses the demonstration into ordered multimodal prompt tokens and uses training-only dual-edge attention supervision to expose its latent stage structure. During execution, an episode-local fast-weight memory causally encodes realized action--observation transitions and modulates prompt cross-attention, producing a progress-aligned context for a unified diffusion action expert without test-time stage labels or stage-specific policies. PACE improves success from 88.9\% to 94.0\% on LIBERO-Gen Goal Chain, from 79.1\% to 83.3\% on Spatial Combination, and from 33.3\% to 73.3\% on the two-step Block Routing tasks. Failure analysis further indicates that structured demonstration alignment and causal execution memory jointly mitigate stage confusion.

\end{abstract}

\section{INTRODUCTION}

Enabling robots to adapt to new tasks from minimal user input without retraining is a central objective of general-purpose robotic manipulation. Recent visuomotor policies have increasingly supported task conditioning through language, goal images, and multimodal demonstrations \cite{octo2024,jain2024vid2robot,jiang2023vima,patel2026bpp}. Among these, a perception--action demonstration conveys the global task objective, the sequence of operations, and fine-grained motion details, enabling policies to adapt to new execution objectives without updating model parameters \cite{duan2017oneshot,james2018taskembedded,devin2019cpv,jain2024vid2robot,xu2023xskill,fu2024icrt,patel2026bpp}.

\begin{figure}[!t]
    \centering
    \includegraphics[width=\columnwidth]{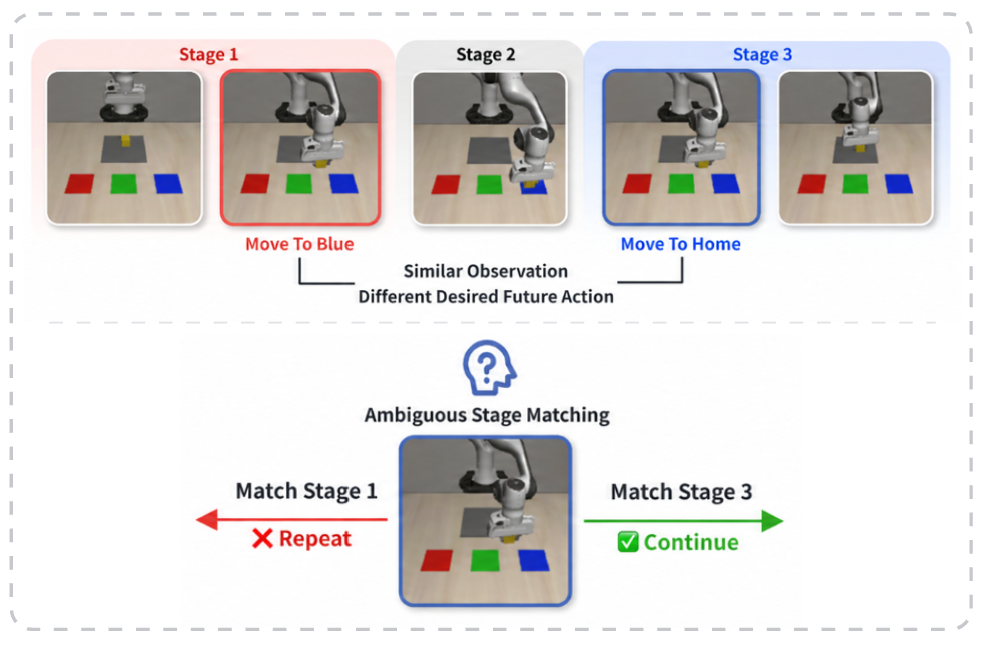}
    \caption{Stage confusion in long-horizon demonstration-conditioned manipulation. Visually similar observations can recur at different stages while requiring different future actions (top). Observation-only matching may therefore retrieve an earlier stage and repeat completed behavior instead of continuing from the current progress (bottom).}
    \label{fig:stage_confusion}
\end{figure}

However, long-horizon multi-stage manipulation requires a policy to continually track task progress and interpret the demonstration relative to the current execution state. This is particularly challenging when different stages contain similar or recurring observations: similar states may require different actions depending on what has already been completed. Retrieving a demonstration segment inconsistent with the current progress may cause the policy to regress to an earlier stage, skip or repeat a stage, or terminate prematurely. We refer to this class of failures as \textbf{stage confusion}, illustrated in Fig.~\ref{fig:stage_confusion}. Stage confusion primarily arises from two related factors: \textbf{cross-stage observation aliasing}, whereby distinct latent stages exhibit similar observations, and \textbf{demonstration--execution temporal-scale mismatch}, whereby the robot and the demonstrator complete the same behavioral stage at different speeds.


Existing approaches address this problem only partially. Explicit stage or subgoal models often require predefined task decompositions, semantic subgoals, or manual stage annotations \cite{sutton1999options,pertsch2021spirl,shiarlis2018taco,kipf2019compile}, while frame- or window-level localization is vulnerable to recurring observations and execution-speed variation \cite{sermanet2018tcn,dwibedi2019tcc}. Long-context and memory-augmented policies retain longer execution histories \cite{dai2019transformerxl,ba2016fastweights,sun2024ttt,jiang2026robottt}, but typically use history as additional input to action prediction rather than to continually revise which part of the demonstration is relevant. These limitations motivate a mechanism that recovers stage structure from demonstrations without manual annotations and uses causal execution history to dynamically determine which part of the demonstration should guide the current behavior.

Event cognition research suggests that humans parse continuous activity into structured events and update an active event model at perceptual boundaries \cite{zacks2001eventstructure,zacks2007event,kurby2008segmentation}. Inspired by this view, we adopt a functional abstraction in which an event model is realized not necessarily through parametrically independent modules, but through context dynamically constructed during execution. Specifically, the policy combines current perception with within-episode execution history to locate and aggregate demonstration segments consistent with the current progress, forming an evolving event context that enables a shared policy to produce stage-appropriate behavior.

Building on this insight, we introduce \textbf{Progress-Aligned Context for Execution (PACE)}, a stateful progress-alignment method for long-horizon demonstration-conditioned manipulation. During training, PACE automatically extracts and monotonically aligns action-stage boundaries, imposing dual-edge attention supervision and effective-support regularization. During execution, an episode-local fast-weight memory causally encodes realized action--observation transitions and modulates observation queries before cross-attention with the full demonstration. The resulting progress-aligned demonstration context conditions a unified diffusion action expert. Stage boundaries are used only for training supervision; at inference time, PACE requires neither explicit event segmentation, discrete stage prediction, nor stage-specific policies.

We evaluate PACE on compositional long-horizon manipulation benchmarks and real-world robotic tasks. On the LIBERO-Gen Spatial Combination and Goal Chain, PACE improves the success rate from 79.1\% to 83.3\% and from 88.9\% to 94.0\%. On the two-step Block Routing tasks, which exhibit pronounced cross-stage observation aliasing, PACE raises the mean success rate from 33.3\% to 73.3\%. Stage-level analyses further show that structured demonstration alignment and causal execution memory work synergistically to suppress stage confusion.

Our main contributions are as follows:
\begin{itemize}
\item We systematically study \textbf{stage confusion} in long-horizon demonstration-conditioned manipulation, attributing it to cross-stage observation aliasing and demonstration--execution temporal-scale mismatch.
\item We introduce \textbf{PACE}, centered on a causally updated, episode-local fast-weight state that encodes realized action--observation transitions and modulates observation queries before prompt cross-attention, allowing visually similar current observations to attend to different demonstration stages under different execution histories. PACE further combines automatic stage-structure discovery with dual-edge attention supervision and effective-support regularization, without requiring manual stage annotations or stage-specific policies.
\item We validate PACE on compositional simulation tasks and real-world robotic tasks with held-out demonstrations, and use stage-level analyses to elucidate how it mitigates stage confusion.
\end{itemize}

\section{RELATED WORK}

\begin{figure*}[!t]
    \centering
    \includegraphics[width=0.9\textwidth]{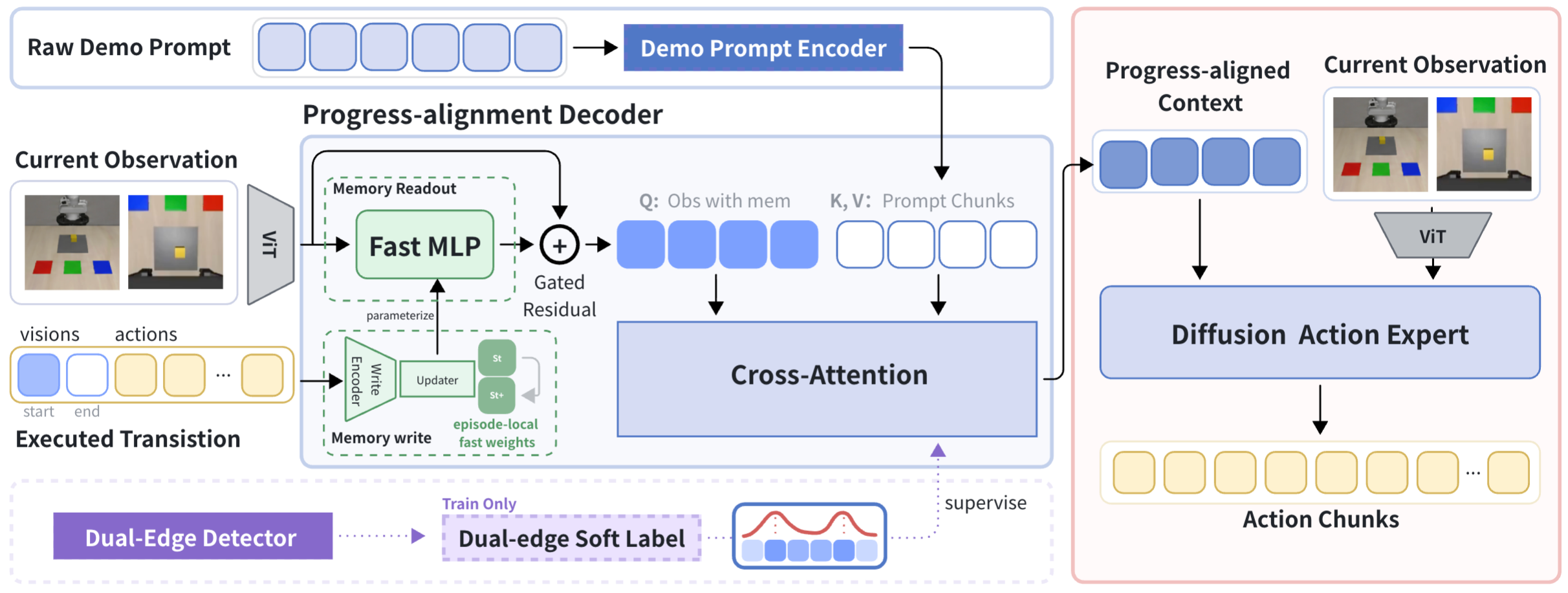}
    \caption{Architecture overview of PACE. The demonstration prompt encoder produces ordered prompt chunks. During execution, episode-local fast weights encode realized action--observation transitions and modulate observation queries before prompt cross-attention. Training-only dual-edge attention supervision is applied to the demonstration attention. The resulting progress-aligned context and current observation condition a diffusion action expert.}
    \label{fig:pace_architecture}
\end{figure*}

\subsection{Demonstration-Conditioned Robot Policies}

Complete perception--action demonstrations provide a natural
inference-time interface for specifying task objectives, operation
sequences, and motion details without updating policy parameters.
Existing methods relate human demonstrations to robot observations
through cross-attention or learned cross-embodiment representations
\cite{jain2024vid2robot,xu2023xskill}, retrieve or align relevant
demonstration segments \cite{papagiannis2024rplusx,kedia2024rhyme},
encode task identity from one or a few demonstrations
\cite{duan2017oneshot,james2018taskembedded}, or jointly model
demonstration and execution trajectories as prompted, in-context
sequences \cite{jiang2023vima,fu2024icrt,patel2026bpp}.
In particular,
Behavior Prompting Policy (BPP) treats a complete perception--action
demonstration as a behavioral prompt for closed-loop action generation
\cite{patel2026bpp}. However, these approaches generally use the
demonstration as a static conditioning sequence and do not continually
reinterpret it according to realized execution progress, leaving them
susceptible to cross-stage observation aliasing.

\subsection{Temporal Alignment and Stage-Structured Manipulation}

Temporal correspondence methods, including frame-level metric learning,
temporal cycle consistency, dynamic time warping, and optimal transport,
align sequences with different appearances or execution speeds
\cite{sermanet2018tcn,dwibedi2019tcc,sakoe1978dtw,cuturi2013sinkhorn}.
Local correspondences become ambiguous when visually similar states
recur, while sequence-level alignments are typically computed offline
or remain fixed during execution. Related hierarchical approaches
represent progress using skills, options, subgoals, or keyframes
\cite{sutton1999options,pertsch2021spirl,shiarlis2018taco,kipf2019compile}, but often require predefined
decompositions, semantic labels, or separate skill policies. Although
unsupervised skill discovery reduces annotation requirements
\cite{eysenbach2019diayn}, it does not directly establish an
execution-conditioned correspondence with a new demonstration.
Latent-plan methods learn reusable long-horizon structure from play or
human video \cite{lynch2020latentplans,wang2023mimicplay}, but likewise
do not continually realign a provided demonstration using the realized
execution history.
PACE instead learns ordered demonstration structure without manual
stage labels and dynamically interprets that structure during execution.

\subsection{History and Memory for Long-Horizon Manipulation}

Long-horizon manipulation is partially observable because similar
current observations may require different actions depending on prior
interactions. Recurrent policies, history-aware Transformers, and
memory-augmented models address this problem by retaining execution
history \cite{dai2019transformerxl,ba2016fastweights}. Test-time
training further represents history as recurrently updated fast weights
\cite{sun2024ttt}, and RoboTTT applies this mechanism to long-horizon
visuomotor control \cite{jiang2026robottt}. These approaches primarily
use history to improve action prediction. In contrast, PACE uses an
episode-local fast-weight state to modulate observation queries before
prompt cross-attention, allowing execution history to determine which
part of the demonstration guides the current action.



\section{PROBLEM STATEMENT}

We consider long-horizon manipulation conditioned on a single perception--action demonstration $\mathcal P=\{(o_j^p,a_j^p)\}_{j=1}^{T_p}$. At execution timestep $t$, the policy predicts an action chunk of horizon $H$ from the current observation and strictly causal episode history:
\begin{equation}
\begin{aligned}
\mathcal H_t
&=
\left\{
\left(o_k^e,a_k^e,o_{k+1}^e\right)
\right\}_{k=1}^{t-1},\\
a^e_{t:t+H-1}
&\sim
\pi_\theta\!\left(
\cdot\mid o_t^e,\mathcal H_t,\mathcal P
\right).
\end{aligned}
\label{eq:prompt_conditioned_policy}
\end{equation}

Here, $o_j^p$ and $a_j^p$ respectively denote the observation and action at demonstration timestep $j$, and $T_p$ is the demonstration length. Each tuple $(o_k^e,a_k^e,o_{k+1}^e)$ denotes a realized action--observation transition: action $a_k^e$ is executed under observation $o_k^e$, producing the subsequent observation $o_{k+1}^e$. At execution timestep $t$, $o_t^e$ denotes the current execution observation, and $\mathcal H_t$ contains all realized transitions prior to action prediction at timestep $t$. The policy $\pi_\theta$, parameterized by $\theta$, predicts the execution action sequence $a^e_{t:t+H-1}$ over a horizon of $H$ steps. The superscripts $p$ and $e$ distinguish prompt demonstration and execution quantities, respectively. The parameters $\theta$ remain fixed at test time; any episode-local state is reset between episodes. The demonstration $\mathcal P$ is the sole behavior specification, with no test-time stage label or progress signal.

The demonstration and execution follow the same high-level operation order but may differ in trajectory length and stage duration. Their latent progress correspondence is represented by
\begin{equation}
\phi:
\{1,\ldots,T_e\}
\rightarrow
\{1,\ldots,T_p\},
\label{eq:latent_progress_mapping}
\end{equation}
where $T_e$ is the execution horizon and $\phi(t)$ maps execution timestep $t$ to its corresponding latent position in the demonstration. The mapping $\phi$ is monotonic but need not advance linearly or at a constant rate. It is latent rather than explicitly predicted, and the relevant guidance may span a contiguous demonstration interval around $\phi(t)$. Given training tasks $\mathcal T_{\mathrm{train}}$, we learn $\pi_\theta$ to maximize complete-task success on evaluation tasks $\mathcal T_{\mathrm{test}}$, including unseen task compositions or held-out demonstrations, without manually annotated stage boundaries or predefined task decompositions.

\section{METHOD}

\subsection{Architecture}

PACE is a demonstration-driven closed-loop method for long-horizon, multi-stage manipulation. Given a complete behavior demonstration and the current observation, PACE continually updates an episode-local state from strictly causal execution history. This state modulates cross-attention between the current observation and the demonstration, dynamically producing a progress-aligned demonstration context that conditions a unified diffusion action expert.

As shown in Fig.~\ref{fig:pace_architecture}, PACE comprises four components: a demonstration prompt encoder, dual-edge attention supervision, a stateful progress-alignment decoder, and a unified diffusion action expert. The prompt encoder represents the ordered demonstration structure, while dual-edge attention supervision shapes this representation during training. The decoder and action expert infer execution progress from realized history and generate closed-loop actions.


\subsection{Demonstration Prompt Encoder}

\begin{figure}[!htbp]
    \centering
    \includegraphics[width=0.9\columnwidth]{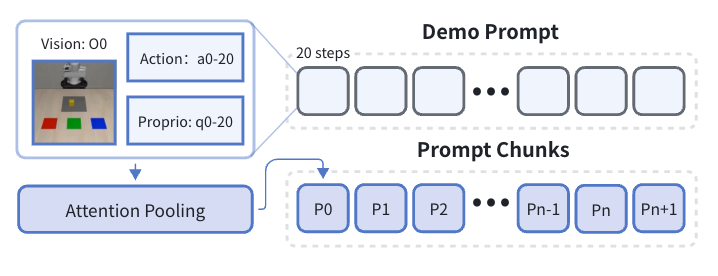}
    \par\vspace{-0.5ex}
    \textbf{(a)}
    \par\vspace{0.8ex}
    \includegraphics[width=0.9\columnwidth]{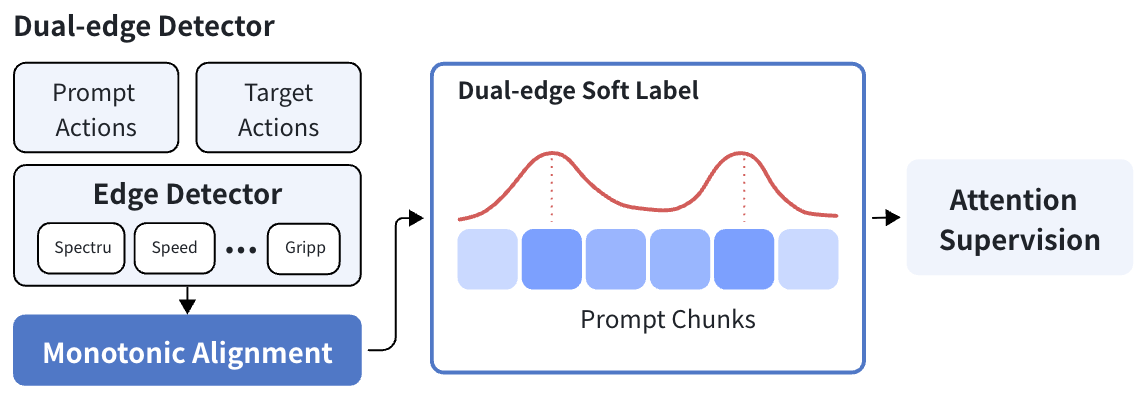}
    \par\vspace{-0.5ex}
    \textbf{(b)}
    \caption{Stage-structure representation in PACE. (a) Each 20-step multimodal demonstration chunk is aggregated by attention pooling into one ordered prompt token. (b) Detected prompt and target boundaries are monotonically aligned to construct a soft dual-edge target over prompt chunks for attention supervision.}
    \label{fig:prompt_structure_modules}
\end{figure}

A long-horizon demonstration typically contains several behavior stages with a clear temporal order. Encoding every frame is computationally expensive and can overemphasize local visual similarity. Drawing on the temporal chunking strategy of BPP \cite{patel2026bpp}, we divide the demonstration into fixed-length chunks of $K=20$ control steps, corresponding to approximately one second at 20~Hz. Actions in an incomplete final chunk are zero-padded and masked. We retain only the first visual frame of each chunk, while preserving its proprioceptive and action sequences, as summarized in Fig.~\ref{fig:prompt_structure_modules}(a).

For chunk $i$, a shared pretrained CLIP ViT \cite{radford2021clip,dosovitskiy2021vit} encodes the first frame from each camera, with its CLS token used as the visual representation. End-effector pose and gripper state are linearly projected to the common dimension $d$, while the $K$ actions are flattened and encoded by an action MLP. Let $\mathcal V_i=\{\mathbf v_i^c\}_{c=1}^{C}$ and $\mathcal P_i=\{\mathbf p_{i,j}\}_{j=0}^{K-1}$ denote the visual and proprioceptive features. These modalities are then fused using attention pooling \cite{vaswani2017attention}, which uses a single learnable latent token as the query and the \(d\)-dimensional visual and proprioceptive tokens, together with the action token \(\mathbf a_i\), as keys and values. Finally, a learnable temporal positional embedding \(\mathbf e_i\) is added to obtain the chunk token:
\begin{equation}
\mathbf z_i=\operatorname{AttnPool}
(\mathcal V_i,\mathcal P_i,\mathbf a_i)+\mathbf e_i .
\label{eq:demo_chunk_token}
\end{equation}
 Here, $i\in\{0,\ldots,N_p-1\}$ indexes demonstration chunks, $N_p=\lceil T_p/K\rceil$ is the number of prompt chunks, and $C$ is the number of camera views. The vector $\mathbf v_i^c$ denotes the visual feature from camera $c$ for chunk $i$, while $\mathbf p_{i,j}$ denotes the proprioceptive feature at within-chunk step $j\in\{0,\ldots,K-1\}$. The action feature $\mathbf a_i$ is obtained by flattening and encoding the $K$ actions in chunk $i$, and $\mathbf e_i\in\mathbb R^d$ is a learnable temporal positional embedding. $\operatorname{AttnPool}(\cdot)$ aggregates the multimodal chunk features into a single $d$-dimensional token $\mathbf z_i$. The resulting demonstration memory $\mathbf Z^p=[\mathbf z_0,\ldots,\mathbf z_{N_p-1}]$ is a continuous, ordered sequence at a fixed temporal resolution rather than a sequence of discrete event labels.

\subsection{Dual-Edge Attention Supervision}

PACE automatically detects stage transitions from both the target execution and prompt demonstration trajectories for dual-edge attention supervision during training. At each timestep, the supervision pipeline compares action patterns in two neighboring windows and combines changes in local action spectrum, motion direction, mean, end-effector speed, pausing behavior, jerk, and gripper state. Continuous motion channels are robustly normalized, while gripper transitions are handled separately as explicit boundary cues. Weighted change signals are filtered using peak height, prominence, and minimum-distance constraints, and the trajectory start is added as a fixed anchor.


Execution and demonstration trajectories may differ in speed, stage duration, and the number of detected boundaries. We therefore use dynamic programming to obtain a monotonic correspondence between their ordered anchor sequences, following the ordering principle of classical sequence alignment \cite{sakoe1978dtw}. The matching cost considers relative temporal position, boundary type, repeated matches, and skipped prompt anchors. This preserves behavior order while tolerating a small number of missing or redundant detections. Figure~\ref{fig:prompt_structure_modules}(b) summarizes the resulting training-only supervision pipeline.

For a training timestep, the correspondence identifies the left and right boundaries, $\mu_L$ and $\mu_R$, of the matching demonstration stage. We convert them into masked Gaussian distributions over prompt chunks with $\sigma=0.75$ and supervise both the aggregated decoder attention and each observation query:
\begin{equation}
\begin{aligned}
\mathbf y_t
&=\tfrac{1}{2}\mathcal G_{\sigma}(\mu_L)
 +\tfrac{1}{2}\mathcal G_{\sigma}(\mu_R),\\
\mathcal L_{\mathrm{edge}}
&=\operatorname{CE}(\mathbf y_t,\bar{\boldsymbol\alpha}_t)
+\gamma\,\mathbb E_q
\left[\operatorname{CE}(\mathbf y_t,\boldsymbol\alpha_{t,q})\right].
\end{aligned}
\label{eq:dual_edge_objective}
\end{equation}
Here, $\mu_L$ and $\mu_R$ denote the prompt-chunk indices of the aligned left and right stage boundaries for training timestep $t$. $\mathcal G_\sigma(\mu)$ is a Gaussian distribution centered at boundary location $\mu$ with standard deviation $\sigma=0.75$, evaluated over valid prompt-chunk indices and renormalized after masking. The target distribution $\mathbf y_t$ is the equally weighted mixture of the two boundary-centered distributions. $\bar{\boldsymbol\alpha}_t$ is the demonstration attention averaged across attention heads and observation queries, while $\boldsymbol\alpha_{t,q}$ is the demonstration-attention distribution produced by observation query $q$. $\operatorname{CE}(\cdot,\cdot)$ denotes cross-entropy, $\mathbb E_q[\cdot]$ averages over all observation queries, and $\gamma$ controls the relative weight of query-level attention supervision. The dual-edge target describes a stage interval rather than a single matching frame, allowing flexible aggregation within the stage and accommodating demonstration--execution speed variation. We omit this supervision for an open final stage without a reliable right boundary.

\subsection{Stateful Progress-Alignment Decoder}

Stage structure alone does not identify the robot's current progress when visually similar states recur across stages. PACE therefore maintains an episode-local fast-weight memory, represented by a fixed $d_m\!\rightarrow\!h_m\!\rightarrow\!d_m$ two-layer MLP $f_{\mathbf s}$. Its state $\mathbf s_k=\{W_1,b_1,W_2,b_2\}$ contains all MLP weights and biases rather than a low-rank subset. Thus, ``fixed-size'' means that the parameterized state has constant size and stores no growing trajectory buffer. Following the TTT-MLP view of a learnable model as recurrent state \cite{sun2024ttt,jiang2026robottt}, PACE updates this state after every completed execution segment of $S=12$ control steps, independently of the $K=20$-step prompt chunks. Each completed segment represents realized action--observation transitions and is encoded from its valid actions and start/end multi-camera observation features; the resulting representation is projected into a normalized key $\mathbf k_k$ and value $\mathbf v_k^{\mathrm{mem}}$.

Each episode starts from a shared learnable initialization $\mathbf s_0$, and each valid write performs a local key--value reconstruction update:
\begin{equation}
\begin{aligned}
\mathcal L_{\mathrm{write}}^k
&=\frac{1}{d_m}
\left\|f_{\mathbf s_k}(\mathbf k_k)
-\mathbf v_k^{\mathrm{mem}}\right\|_2^2,\\
\mathbf s_{k+1}
&=\mathbf s_k-\eta\nabla_{\mathbf s_k}
\mathcal L_{\mathrm{write}}^k .
\end{aligned}
\label{eq:fast_state_update}
\end{equation}
Here, $k$ indexes completed segments, $d_m$ is the memory dimension, and $\eta$ is a learnable inner-update step size. The reconstruction objective defines only the state transition and is not added to the outer loss; the shared initialization, projections, and $\eta$ are learned by the outer control objectives. Training backpropagates through at most 16 consecutive writes and applies stop-gradient between truncated-BPTT windows. Writes are strictly causal: only segments preceding the prediction time are used in training, and only realized action--observation transitions are written during deployment. With $d_m=64$ and $h_m=128$, the state contains 16,576 scalars and each core fast-MLP write requires approximately $5d_mh_m\approx4.1\times10^4$ multiply--accumulate operations, excluding segment encoding; this cost does not grow with episode length.

Each decoder layer applies observation self-attention, injects a gated fast-memory readout, and only then cross-attends to the complete demonstration:
{\small
\begin{equation}
\begin{gathered}
\Delta\mathbf Q_t^l
=W_o^l f_{\mathbf s_{k(t)}}\!\left(
W_r^l\operatorname{LN}(\widetilde{\mathbf Q}_t^l)\right),\\
\widehat{\mathbf Q}_t^l
=\widetilde{\mathbf Q}_t^l+\tanh(g_l)\Delta\mathbf Q_t^l,\quad
\mathbf C_t^l
=\mathrm{MHA}(\widehat{\mathbf Q}_t^l,\mathbf Z^p,\mathbf Z^p).
\end{gathered}
\label{eq:state_conditioned_attention}
\end{equation}
}
Here, \(\operatorname{LN}\) denotes layer normalization, and \(\operatorname{MHA}\) denotes multi-head attention. $k(t)$ denotes the most recently completed execution segment at timestep $t$, $\widetilde{\mathbf Q}_t^l$ is the observation representation after self-attention, $W_r^l$ and $W_o^l$ project into and out of the memory space, and $g_l$ gates the history-dependent correction. The modulated observation tokens $\widehat{\mathbf Q}_t^l$ query the complete demonstration memory $\mathbf Z^p$, which supplies keys and values. Because memory modulation precedes prompt cross-attention, the same observation can retrieve different demonstration stages under different execution histories. The final context $\mathbf C_t=\mathbf C_t^L$ is concatenated with
the original observation tokens to condition action generation.


The progress-alignment decoder component in Figure~\ref{fig:pace_architecture} summarizes the causal memory write, fast-weight readout, and history-modulated prompt cross-attention.

\subsection{Unified Diffusion Action Expert}

PACE uses a single conditional 1D U-Net across tasks and behavior stages rather than maintaining stage-specific policies. Task semantics and current progress are specified implicitly by the current observation and progress-aligned demonstration context. We train the expert with the standard diffusion denoising objective \cite{ho2020ddpm,chi2023diffusionpolicy} and jointly apply the dual-edge attention supervision loss to its final-layer demonstration attention. Entropy-based effective-support regularization further prevents attention from spreading uniformly across the complete demonstration while preserving flexible weighting within the relevant interval. The complete objective is
\begin{equation}
\mathcal L=\mathcal L_{\mathrm{diff}}
+\lambda_{\mathrm{edge}}(u)\mathcal L_{\mathrm{edge}}
+\lambda_{\mathrm{sup}}\mathcal L_{\mathrm{sup}} .
\label{eq:pace_objective}
\end{equation}
Here, $\mathcal L_{\mathrm{diff}}$ is the standard diffusion denoising loss, $\mathcal L_{\mathrm{edge}}$ is the dual-edge attention supervision loss in Eq.~\eqref{eq:dual_edge_objective}, and $\mathcal L_{\mathrm{sup}}$ is the effective-support regularization term on demonstration attention. The scalar $u$ denotes training progress, $\lambda_{\mathrm{edge}}(u)$ is a schedule-dependent weight that is gradually increased during training, and $\lambda_{\mathrm{sup}}$ is the fixed weight of the effective-support regularization. Thus, the action objective remains shared across all stages while the auxiliary terms shape how the demonstration is read.

At inference time, dual-edge attention supervision is removed entirely; no explicit event segmentation or discrete stage prediction is required. PACE updates its fast state from the current observation and realized execution history, constructs a continuous progress-aligned context, executes part of the action sequence produced by the diffusion expert, and replans after incorporating the new outcome. This yields closed-loop progress alignment and action generation.

\section{EXPERIMENTS}

This section evaluates the overall closed-loop performance of PACE on long-horizon, multi-stage demonstration-conditioned manipulation tasks, using complete-task success as the primary criterion. We defer analysis of stage confusion, demonstration structure, and the progress-alignment mechanism to Sec.~\ref{sec:ablation}.

\subsection{Experimental Setup}

\begin{figure}[!t]
    \centering
    \includegraphics[width=\columnwidth]{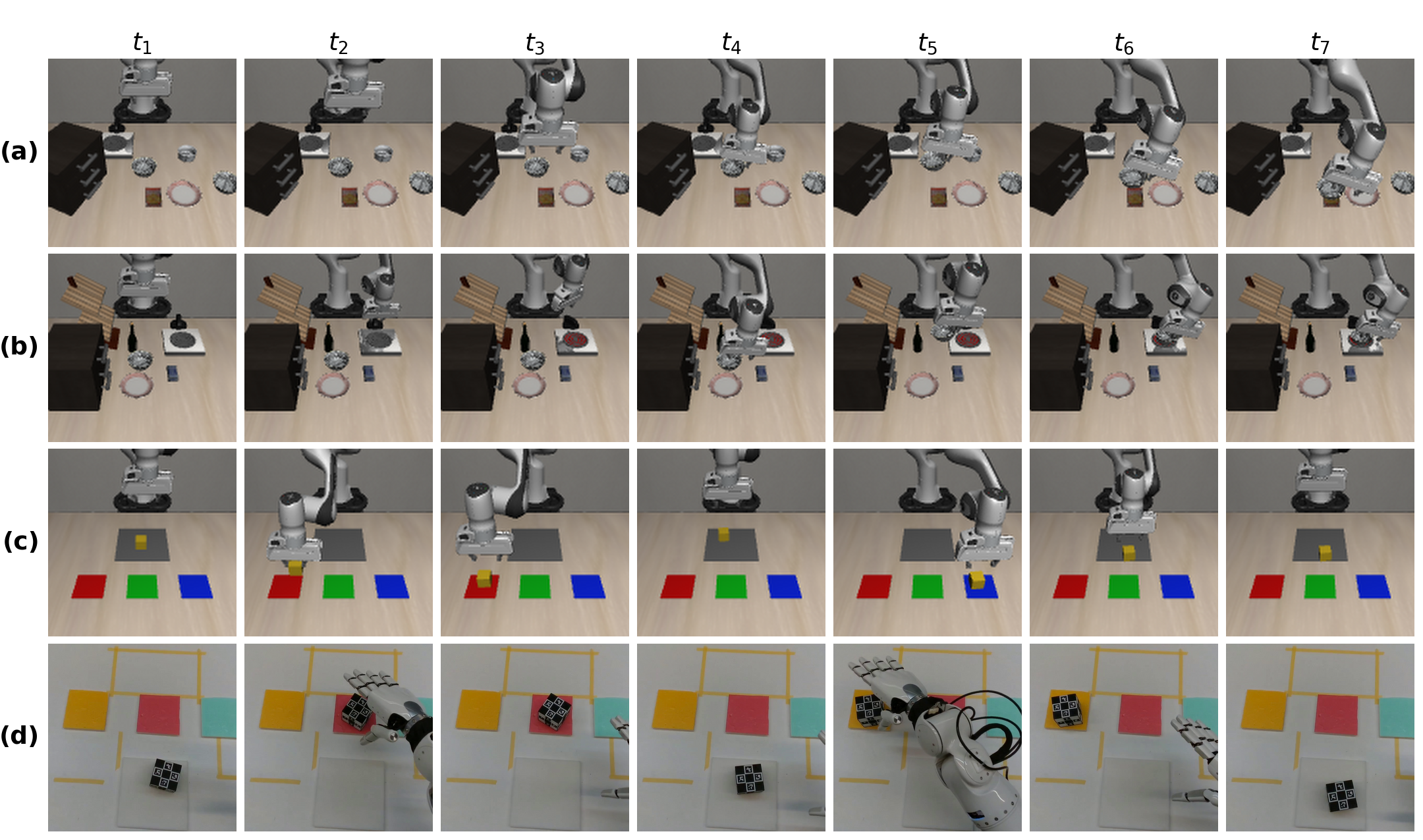}
    \caption{Representative execution sequences: (a) simulation Spatial Combination, (b) simulation Goal Chain, (c) simulation Block Routing, and (d) real-world Block Routing.}
    \label{fig:simulation_tasks}
    \label{fig:real_world_tasks}
\end{figure}

\textbf{Simulation tasks.} We evaluate three simulation task families with different stage compositions and execution structures, illustrated in Fig.~\ref{fig:simulation_tasks}. \textit{LIBERO-Gen Goal Chain} concatenates a prerequisite interaction, such as opening a drawer, with a subsequent object-centric goal, testing unseen compositions of goals and behavior chains. \textit{LIBERO-Gen Spatial Combination} recombines manipulated objects and target locations from LIBERO-Spatial \cite{liu2023libero} to evaluate demonstration-conditioned generalization to unseen spatial relations; both LIBERO-Gen suites follow BPP \cite{patel2026bpp}. \textit{Block Routing} repeatedly moves the same block among red, green, and blue target regions, returning it to a common home region between consecutive operations. It contains three single-region routes and nine ordered two-region routes, for 12 tasks in total, and therefore presents recurring observations across behavior stages. Together, these tasks cover the two main challenges considered by PACE: cross-stage observation aliasing and demonstration--execution temporal-scale mismatch.

\textbf{Real-robot tasks.} We further evaluate PACE on real-robot Block Routing under visual perturbations, action errors, and stochastic interaction outcomes, as illustrated in Fig.~\ref{fig:real_world_tasks}. The task mirrors the simulation setup and allows us to examine behavioral consistency across simulation and reality.

\begin{figure*}[!t]
    \centering
    \begin{minipage}[t]{0.32\textwidth}
        \centering
        \includegraphics[width=\linewidth]{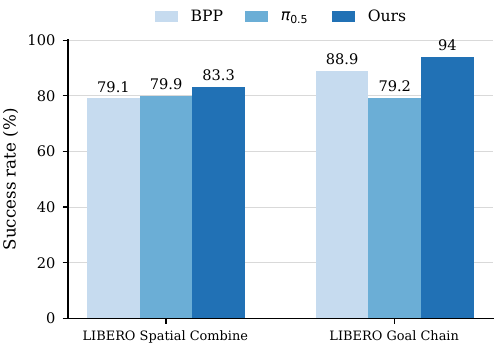}
        \par\textbf{(a)}
    \end{minipage}\hfill
    \begin{minipage}[t]{0.32\textwidth}
        \centering
        \includegraphics[width=\linewidth]{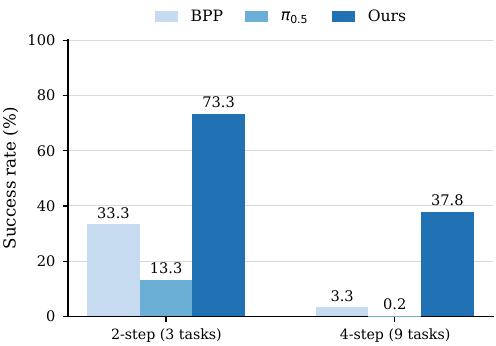}
        \par\textbf{(b)}
    \end{minipage}\hfill
    \begin{minipage}[t]{0.32\textwidth}
        \centering
        \includegraphics[width=\linewidth]{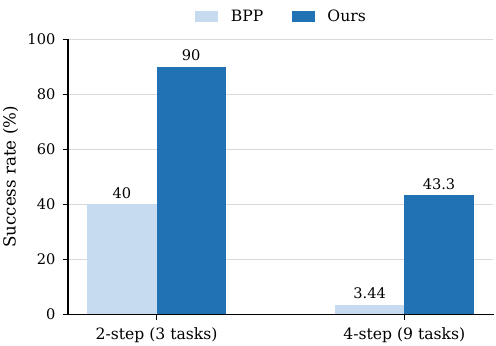}
        \par\textbf{(c)}
    \end{minipage}
    \caption{Complete-task success rates. The horizontal axis denotes the task category, and the vertical axis reports the mean complete-task success rate. \textbf{(a) Simulation Experiment 1:} LIBERO-Gen Spatial Combination and LIBERO-Gen Goal Chain each contain 10 tasks whose stage compositions are unseen during training. Each task is evaluated with 10 rollouts, and the success rate is averaged across tasks. \textbf{(b) Simulation Experiment 2:} Block Routing consists of 2-step and 4-step tasks, with 10 training demonstrations for each task. A 2-step task moves the block to one of the red, green, or blue target regions and then returns it to the home region, yielding three tasks. A 4-step task concatenates an ordered pair of 2-step tasks---for example, $\mathrm{Home}\!\rightarrow\!\mathrm{Red}\!\rightarrow\!\mathrm{Home}\!\rightarrow\!\mathrm{Green}\!\rightarrow\!\mathrm{Home}$---yielding $3\!\times\!3=9$ tasks. Each task is evaluated with 10 rollouts, and the mean success rate is reported for each category. \textbf{(c) Real-robot Block Routing:} The task follows the same setup as (b), using 15 training demonstrations and 10 evaluation rollouts per task; the mean success rate is reported for each category.}
    \label{fig:overall_success_results}
\end{figure*}

\textbf{Baselines.} Behavior Prompting Policy (BPP) \cite{patel2026bpp} is our principal baseline and uses the same form of complete behavior demonstration and current observation as PACE. We additionally compare with $\pi_{0.5}$ \cite{black2025pi05} in simulation. BPP and PACE use the same training split and checkpoint-selection protocol. All methods are evaluated with the same test demonstrations, success criteria, and rollout protocol.

\textbf{Evaluation metric.} Our primary metric is episode-level task success rate. An episode is successful only if every operation stage is completed in the correct order. We evaluate each task with multiple independent demonstrations and rollout seeds and report the mean over all test episodes. No stage label, task-progress signal, or semantic subgoal is provided at test time.

\subsection{Simulation Results}

Figure~\ref{fig:overall_success_results}(a)--(b) summarizes complete-task success rates on the three simulation task families.

On the unseen compositions in Fig.~\ref{fig:overall_success_results}(a), PACE achieves 83.3\% on Spatial Combination and 94.0\% on Goal Chain, improving over BPP by 4.2 and 5.1 percentage points and over $\pi_{0.5}$ by 3.4 and 14.8 points, respectively. On the seen Block Routing tasks in Fig.~\ref{fig:overall_success_results}(b), PACE reaches 73.3\% on 2-step tasks and 37.8\% on 4-step tasks. These results exceed BPP by 40.0 and 34.5 points and $\pi_{0.5}$ by 60.0 and 37.6 points, respectively. The larger gains on Block Routing, particularly when BPP and $\pi_{0.5}$ fall to 3.3\% and 0.2\% on 4-step tasks, show that PACE better preserves complete-task performance as stage dependencies and recurring states increase.

PACE receives neither test-time stage labels nor stage-specific action experts. Its gains therefore arise from progress-dependent interpretation of the same demonstration condition rather than additional task-semantic supervision.

\subsection{Real-Robot Results}

Real-world execution exhibits greater variation in action duration, contact error, and visual appearance than simulation, so the time required to complete a behavior stage often differs from the demonstration. Figure~\ref{fig:overall_success_results}(c) reports the corresponding evaluation.

PACE achieves 90.0\% success on the real-robot 2-step tasks, compared with 40.0\% for BPP, an improvement of 50.0 percentage points. On the more demanding 4-step tasks, PACE reaches 43.3\%, compared with 3.44\% for BPP, improving success by 39.9 percentage points.

\section{ABLATIONS}
\label{sec:ablation}

This section addresses three questions: (1) whether PACE's gains primarily arise from reducing stage confusion; (2) whether demonstration stage structure and stateful progress alignment are both required; and (3) whether the learned demonstration attention dynamically tracks execution progress within the demonstration.

\subsection{Does PACE Reduce Stage Confusion?}

\begin{figure}[!htbp]
    \centering
    \includegraphics[width=0.8\columnwidth]{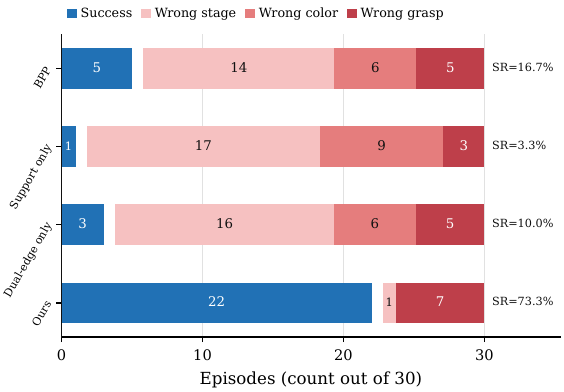}
    \caption{Complete outcome composition on the two-step Block Routing task over 30 rollouts. Each horizontal bar partitions 30 episodes into successes, wrong-stage failures, wrong-color failures, and wrong-grasp failures. Failure subsegments are uniformly compressed to retain the common 30-episode endpoint, while labels report exact counts. The corresponding success rate is reported at right.}
    \label{fig:block3_ablation_outcomes}
\end{figure}

We analyze this issue on the two-step Block Routing task, where the robot will repeatedly pass through the same region, creating distinct stages with highly similar observations that can lead to stage confusion. Here, we compare BPP, PACE, and two ablated variants of PACE. For each method, we conduct 10 rollouts on each task, resulting in 30 evaluation rollouts per method. Figure~\ref{fig:block3_ablation_outcomes} reports the success rate and failure-mode composition of each method. 

We categorize failures into three types: \emph{wrong-stage}, where the policy executes behavior associated with an incorrect stage of the demonstrated sequence; \emph{wrong-color}, where the policy proceeds with the correct stage but selects an incorrect target color; and \emph{wrong-grasp}, where the intended stage and target are correct but execution fails due to low-level grasping or placement errors.

In BPP, wrong-stage failures occur 14 times, accounting for 46.7\% of all trials, whereas PACE produces only one such failure, accounting for 3.3\%. Among failed rollouts, the proportion of wrong-stage failures decreases from 56.0\% in BPP to 12.5\% in PACE. In both the Effective-support regularization only and Dual-edge attention supervision only variants, wrong-stage failures remain the dominant failure mode as in BPP. Together with the 56.6-percentage-point success-rate improvement from BPP to PACE in Fig.~\ref{fig:block3_ablation_outcomes}, these results support the hypothesis that PACE improves long-horizon manipulation primarily by suppressing stage confusion, with most of this gain attributable to the fast-weight memory (FWM).

\subsection{Role of Stage Structure and Stateful Progress Alignment}

Figure~\ref{fig:block3_ablation_outcomes} separates the stage-structure losses (effective-support regularization and dual-edge attention supervision) from the episodic fast-weight state (FWM). Neither structural loss improves success in isolation, while the full model substantially outperforms BPP. This indicates that stage supervision alone cannot resolve recurring observations; it must interact with an execution state that changes demonstration retrieval as progress evolves.

\subsection{Does Demonstration Attention Track Execution Progress?}

\begin{figure}[!htbp]
    \centering
    \makebox[\columnwidth][l]{%
        \hspace*{-2mm}%
        \includegraphics[width=\columnwidth]{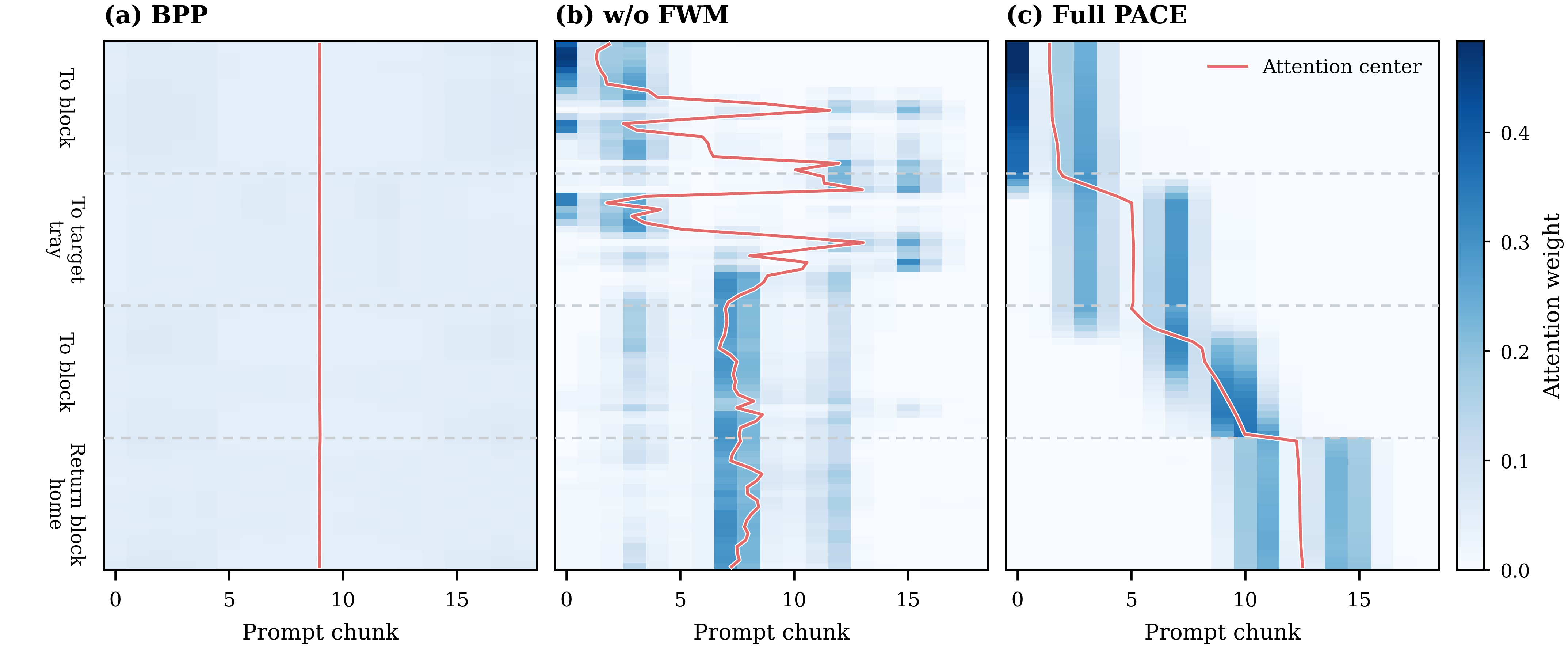}%
    }
    \caption{Attention maps for two-step Block Routing: (a) BPP and (b) PACE w/o FWM show representative failures, while (c) Full PACE shows a representative success. Rows are aligned to four intended stages---\emph{To block}, \emph{To target tray}, \emph{To block}, and \emph{Return block home}. Blue intensity represents normalized attention weight, gray dashed lines separate intervals, and red curves mark the attention center of mass.}
    \label{fig:attention_stage_tracking}
\end{figure}

Figure~\ref{fig:attention_stage_tracking} compares BPP, PACE without fast-weight memory (FWM), and full PACE on the same two-step Block Routing task, prompt demonstration, training seed, and rollout seed. BPP and PACE without FWM produce representative failures, whereas full PACE succeeds. BPP attention remains diffuse and nearly stationary. Without FWM, the dual-edge structure emerges, but attention repeatedly switches between stages under visually aliased observations. In contrast, full PACE maintains an approximately monotonic attention trajectory aligned with execution progress, suggesting that FWM helps prevent retrieval of visually compatible but stage-inappropriate demonstration regions.

To isolate the effect of execution history, we hold the aliased current observation \(X^\ast\) and demonstration \(P\) fixed while varying only the executed episode prefix \(\mathcal H^{(r)}\):
\begin{equation}
\mathbf s^{(r)}=\operatorname{Update}(\mathbf s_0,\mathcal H^{(r)}),
\hspace{0.5em}
\boldsymbol\alpha^{(r)}
=\boldsymbol\alpha(P\mid X^\ast,\mathbf s^{(r)}).
\label{eq:counterfactual_state}
\end{equation}
Here, \(\operatorname{Update}\) applies the causal memory writes in Eq.~\eqref{eq:fast_state_update}, and \(\boldsymbol\alpha^{(r)}\) is the resulting demonstration-attention distribution. We compare matched, earlier-stage, later-stage, shuffled, and reset histories.

\begin{figure}[!htbp]
    \centering
    \includegraphics[width=0.7\columnwidth]{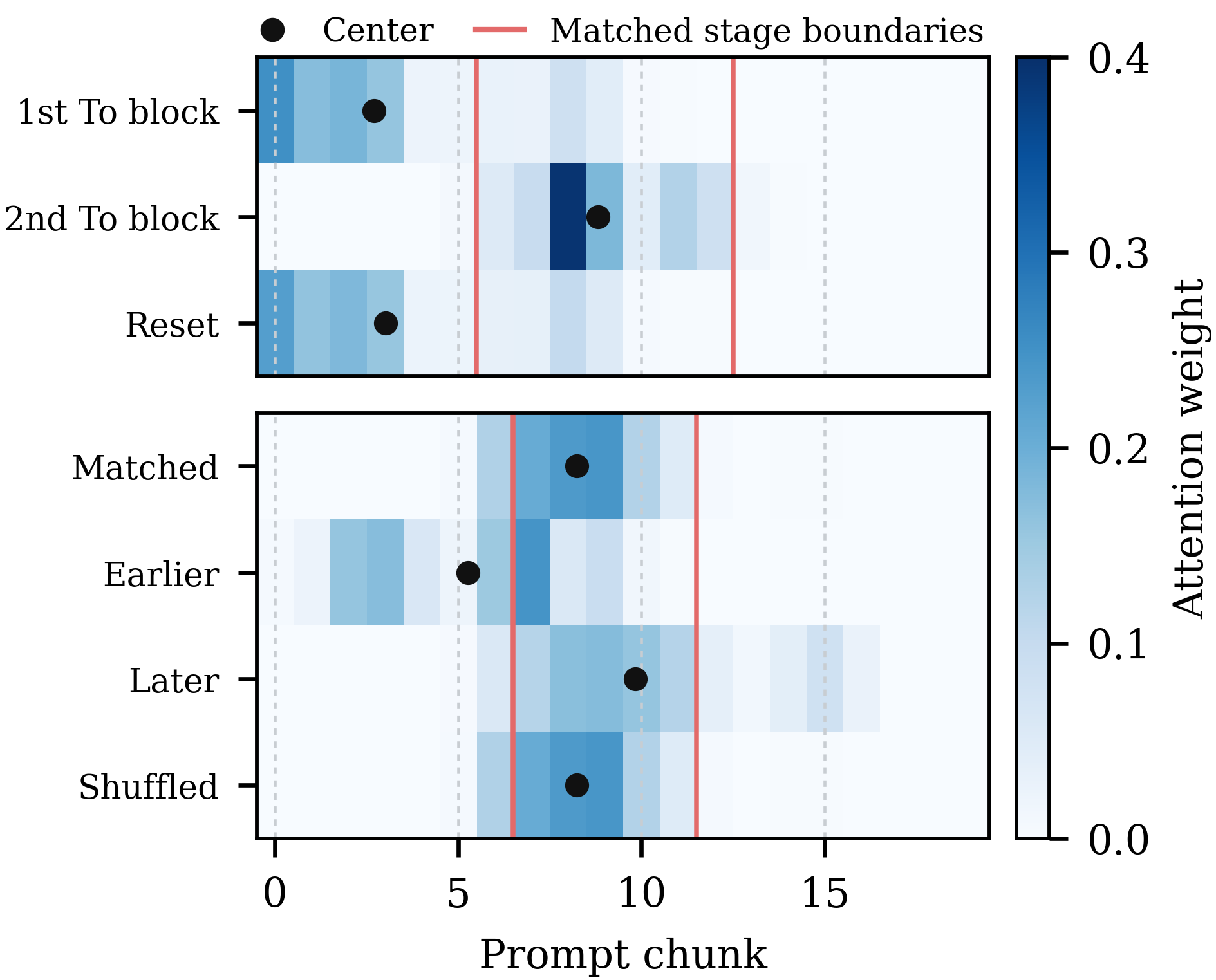}
    \vspace{-2mm}
    \caption{Counterfactual history conditioning with the current observation and demonstration held fixed. The upper panel directly contrasts the visually aliased first and second \emph{To block} occurrences. The lower panel compares matched, earlier, later, and shuffled histories for a randomly selected sample. Black markers denote the attention center of mass, and red lines denote the matched-stage boundaries.}
    \label{fig:counterfactual_stage}
\end{figure}

Figure~\ref{fig:counterfactual_stage} provides qualitative evidence that demonstration retrieval depends not only on the current observation but also on the causal execution history. In the upper panel, we fix the current observation from the first \emph{To block} stage, which is visually similar to observations from the second \emph{To block} stage. Replacing the corresponding fast-state history with the execution history of the second occurrence shifts the attention center from the early prompt chunks to the later matched stage, whereas resetting the fast state moves it back toward the earlier stage. The lower panel, based on a randomly selected current stage, exhibits the same history-dependent behavior: the matched history concentrates attention within the corresponding stage boundaries, while earlier- and later-stage histories shift the attention center toward their respective demonstration regions. Although the shuffled history retains a similar attention center in this example, its overall attention distribution changes. These results indicate that the episode-local fast-weight state makes demonstration retrieval progress-dependent, allowing the same current observation to attend to different demonstration stages under different realized execution histories.

\section{CONCLUSION AND LIMITATIONS}

We presented PACE, a demonstration-driven closed-loop method for long-horizon, multi-stage manipulation. PACE represents a complete demonstration as an ordered multimodal prompt, learns its stage structure through training-only dual-edge attention supervision, and uses a causal episode-local fast-weight memory to continually align demonstration attention with realized execution progress. A unified diffusion action expert then generates actions from the aligned context without explicit stage labels or stage-specific policies. Results on compositional simulation tasks and Block Routing show improved complete-task success and a substantial reduction in wrong-stage failures, supporting the joint role of structured demonstration supervision and stateful progress alignment.

\textbf{Limitations.} The current evaluation covers a limited set of task families and relatively small rollout counts, so broader validation across embodiments, environments, and longer behavior chains is needed. Dual-edge attention supervision relies on action-change cues and assumes a monotonic correspondence between execution and demonstration stages, which may be restrictive for tasks with optional, repeated, or reordered subgoals. Fixed-duration prompt chunks can also blur transitions shorter than the chosen temporal resolution, while fast-weight adaptation introduces additional state-management and computational overhead. Future work should learn boundary discovery end to end, support non-monotonic task structures, and evaluate adaptive temporal resolutions and more efficient episodic memory updates.


\bibliographystyle{IEEEtran}
\bibliography{references}

\end{document}